\documentclass[10pt,twocolumn,letterpaper]{article}

\usepackage[pagenumbers]{cvpr} 

\definecolor{cvprblue}{rgb}{0.21,0.49,0.74}
\usepackage[pagebackref,breaklinks,colorlinks,allcolors=cvprblue]{hyperref}

\usepackage{enumitem}

\usepackage{stfloats}

\usepackage[all]{nowidow}

\usepackage{xcolor}

\def\paperID{16190} 
\def\confName{CVPR}
\def\confYear{2025}

\title{\fontsize{14.25pt}{9pt}\selectfont\textbf{
  SimpleDepthPose: Fast and Reliable Human Pose Estimation with RGBD-Images}}

\author{Daniel Bermuth\\
ISSE\\
University of Augsburg, Germany\\
{\tt\small daniel.bermuth@uni-a.de}
\and
Alexander Poeppel\\
ISSE\\
University of Augsburg\\
{\tt\small poeppel@isse.de}
\and
Wolfgang Reif\\
ISSE\\
University of Augsburg\\
{\tt\small reif@isse.de}
}

\begin{document}
\maketitle

\begin{abstract}
    In the rapidly advancing domain of computer vision, accurately estimating the poses of multiple individuals from various viewpoints remains a significant challenge, especially when reliability is a key requirement. This paper introduces a novel algorithm that excels in multi-view, multi-person pose estimation by incorporating depth information. An extensive evaluation demonstrates that the proposed algorithm not only generalizes well to unseen datasets, and shows a fast runtime performance, but also is adaptable to different keypoints. To support further research, all of the work is publicly accessible.
\end{abstract}

\vspace{-12pt}
\section{Introduction}
\label{sec:intro}
\vspace{-3pt}

In many human-centric applications, determining the precise location and pose of individuals is crucial. Pose estimation, which typically involves identifying the positions of a person's joints, is therefore essential for tasks ranging from motion capture to human-computer interaction.

Traditional methods for pose estimation often rely on markers attached to the body, which can be tracked by specialized cameras. While this approach yields high accuracy, it is also cumbersome as it requires individuals to wear specific clothes, which may not be practical or not possible, for example in public settings or in an operating room.
Marker-less methods, on the other hand, offer greater convenience since they can extract the poses directly from images. However, they face greater computational challenges due to the need to accurately interpret the images.

Using multiple cameras to capture scenes from various angles enhances robustness against occlusions and improves accuracy. Besides using standard RGB cameras, depth cameras can also be employed to provide additional information that can enhance pose estimation.

\begin{figure}[htbp]
  \centering
  \begin{subfigure}{0.9\linewidth}
    \centering
    \includegraphics[width=0.95\linewidth]{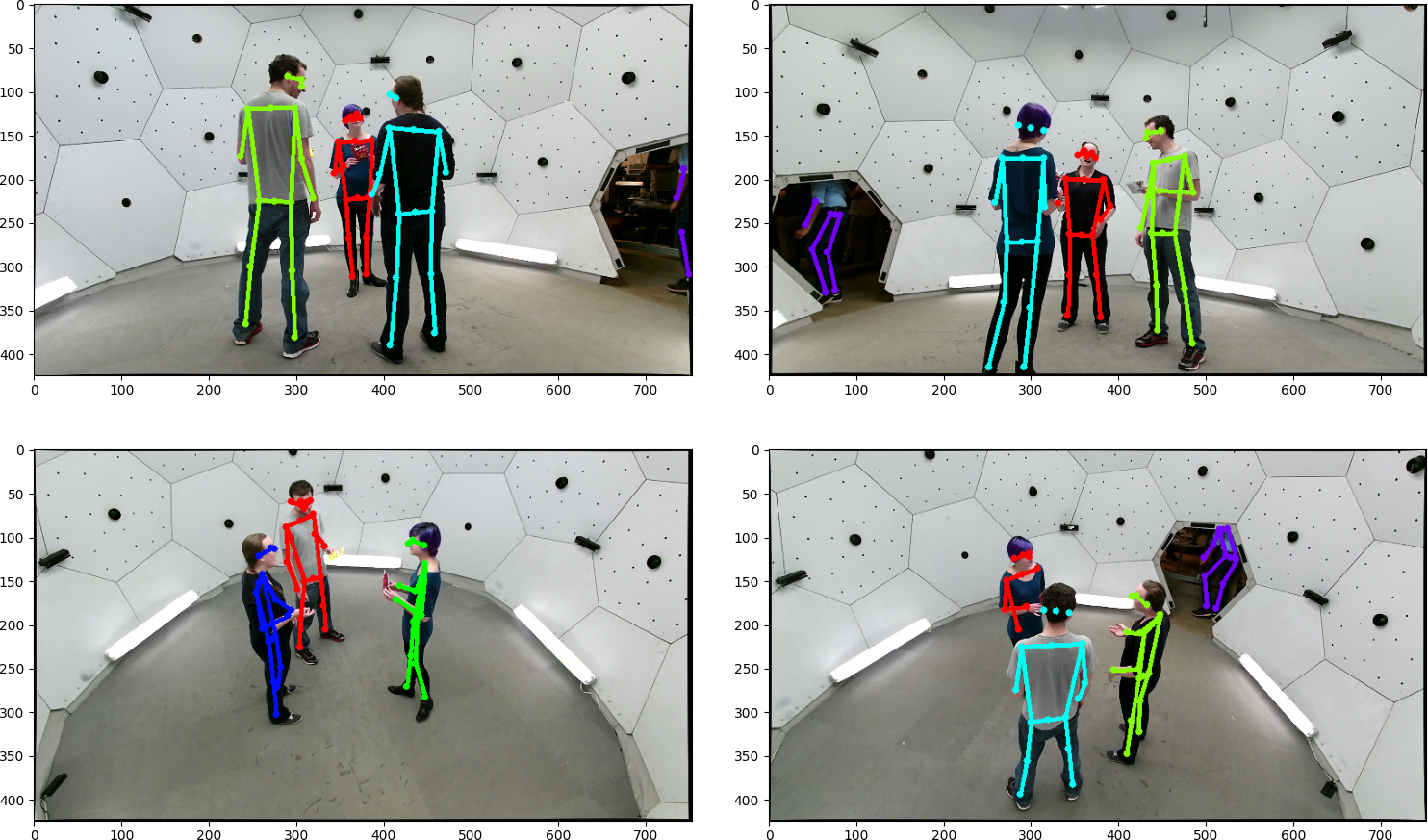}
  \end{subfigure}
  \hfill
  \vspace{6pt}
  \begin{subfigure}{0.9\linewidth}
    \centering
    \includegraphics[width=0.95\linewidth]{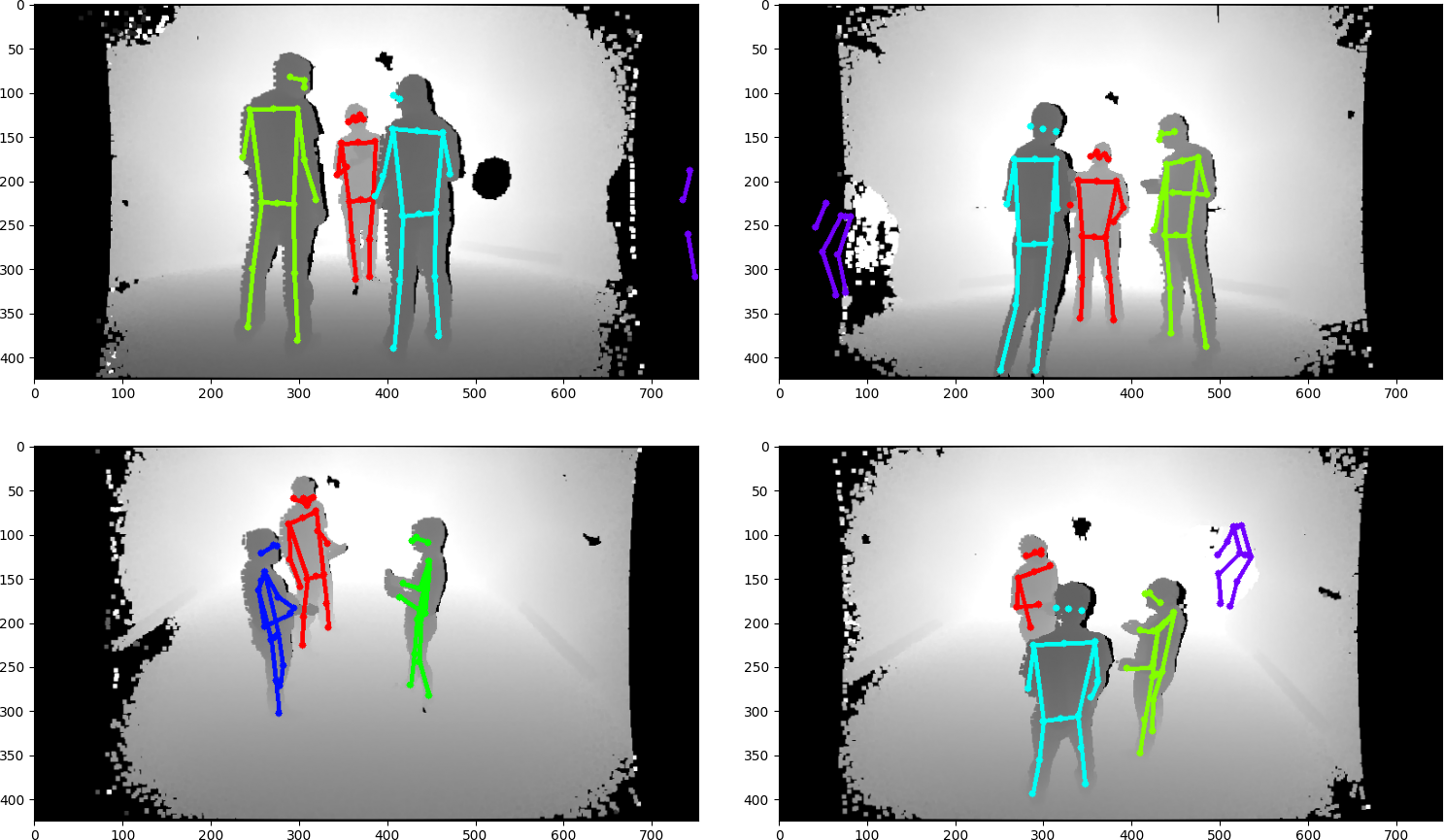}
  \end{subfigure}
  \hfill
  \vspace{6pt}
  \begin{subfigure}{0.89\linewidth}
    \centering
    \includegraphics[width=0.95\linewidth]{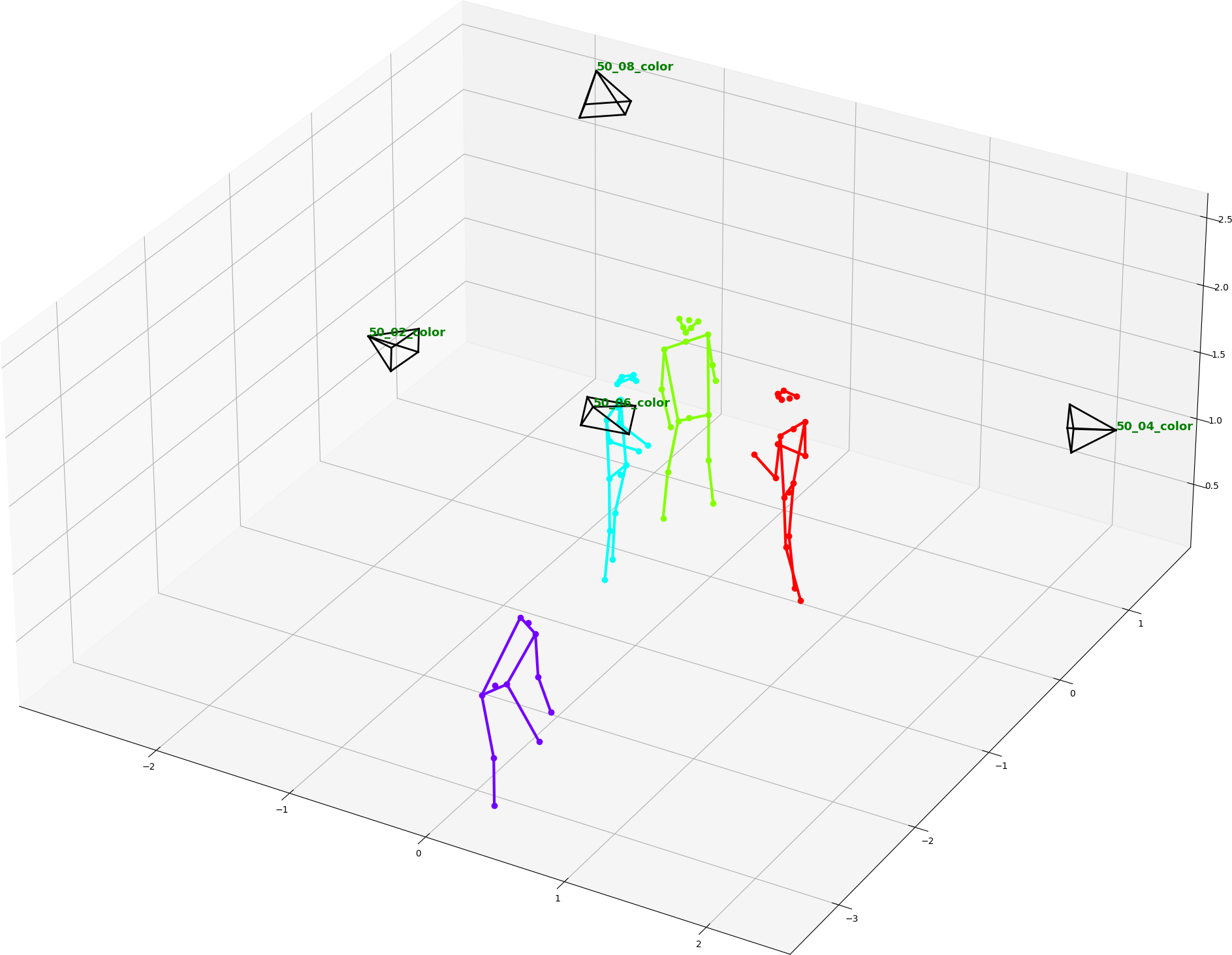}
  \end{subfigure}
  \caption{Example of a multi-person pose estimation from multiple camera views (from the \textit{panoptic} dataset~\cite{joo2015panoptic}). Using the 2D pose detections from the color images (top), a distance to the cameras is extracted from the aligned depth images (center), and the resulting 3D poses of each view are filtered and merged into a final result (bottom).}
  \label{fig:init_example}
\end{figure}

This work presents a simple but fast and reliable algorithm to detect the joints of multiple humans using RGBD-images from multiple views. It does not require additional training and generalizes well across different scenarios.

The source-code of the presented method can be found at: \url{https://gitlab.com/Percipiote/}


\section{Related Work}
\label{sec:relwork}
\vspace{-3pt}

Traditionally, human pose estimation is addressed by a two-phase method. Initially, 2D poses are derived from each image, and subsequently, these are fused to estimate 3D poses. The algorithms can be categorized based on their usage of algorithmic strategies versus learning-based methods.

\vspace{3pt}
From a learning-based perspective, \textit{VoxelPose}~\cite{voxelpose} was one of the first concepts, extending the work of \textit{Iskakov et al.}~\cite{iskakov2019learnable} to multi-person estimations. It projects the joint heatmaps from the 2D images into 3D voxelized space and then estimates a rough center for each person, which is then used to create and extract a concentrated cube around each individual. Following this, the locations of the joints are computed using a second neural network.
\textit{Faster-VoxelPose}~\cite{fastervoxelpose} refined this approach by restructuring the 3D voxel space into multiple 2D and 1D projections to improve efficiency.
\textit{TEMPO}~\cite{choudhury2023tempo} and \textit{TesseTrack}~\cite{reddy2021tessetrack} introduced a temporal dimension to the voxel space to track the poses across frames.
Other methods like \textit{PRGnet}~\cite{wu2021graph} use a graph-based method or directly regress the 3D poses from the 2D features, as in \textit{MvP}~\cite{wang2021mvp}.
\textit{SelfPose3d}~\cite{Srivastav_2024_CVPR} is a recent approach that uses self-supervised training. It adopts the structure of \textit{VoxelPose}, and trains both the 2D and 3D networks with randomly augmented 2D poses.

In terms of algorithmic methods, \textit{mvpose}~\cite{dong2019fast} addresses the problem in two phases: initially, it identifies matching 2D poses across images based on geometric and visual similarities, and then it triangulates these poses to construct the final output.
\textit{mv3dpose}~\cite{tanke2019iterative} employs a graph-matching strategy to allocate poses through epipolar geometry and integrates temporal data to compensate for any missing joint information.
\textit{PartAwarePose}~\cite{chu2021part} speeds up the pose-matching process by utilizing poses from the previous frame, and applies a joint-based filter to correct keypoint inaccuracies caused by occlusions.
\textit{VoxelKeypointFusion}~\cite{voxkeyfuse} uses a voxel-based triangulation concept to predict 3D joint proposals from overlapping views and then uses their reprojections to assign them to persons in those views before grouping them into a final result.

\vspace{6pt}
Since some cameras are capable of capturing depth data in addition to color images, incorporating this depth information could potentially enhance the accuracy of the pose predictions. A few algorithms have already been developed to leverage this additional information.

\textit{OpenPTrack}~\cite{munaro2014openptrack, carraro2019real}, which is frequently employed in robotics, initially calculates the 2D keypoints for each image, and then leverages depth images to determine the distance of each joint to the cameras. It generates a 3D person proposal from each view, which is converted into global world coordinates. Subsequently, these proposals are associated with specific individuals, and a \textit{Kalman-Filter} is applied for joint filtering and temporal smoothing to refine the results.
\textit{MVDeep3DPS}~\cite{kadkhodamohammadi2017multi} utilizes a trainable filter to eliminate inaccurate person proposals before combining them in 3D space. After merging, the method refines these proposals by a calculated confidence score for each body part.
\textit{Ryselis et al.}~\cite{ryselis2020multiple} employed a straightforward strategy of just averaging the 3D poses from the different views.
\textit{Hansen et al.}\cite{hansen2019fusing} generated keypoint heatmaps from depth images and utilized a point cloud to estimate each person's center. They then projected these heatmaps and depth data into a voxelized space to create a 3D pose using a \textit{V2V}\cite{moon2018v2v} network architecture, similar to that of \textit{VoxelPose}. Their source code is not available. 
\textit{PointVoxel}~\cite{pan2023pointvoxel} is a recent work that adopts a similar concept but distinguishes itself by using two separate \textit{V2V}-branches for keypoint and depth voxel-maps instead of merging them directly. It then combines the outputs from these branches. Additionally, it features a synthetic data generator to facilitate generalization across different setups. The source code for this method was not available at the time of writing.
\textit{VoxelKeypointFusion}~\cite{voxkeyfuse} includes a simple voxel-based depth masking approach to remove voxel projections that are not visible in the depth images. 


\section{SimpleDepthPose}
\label{sec:algs1}
\vspace{-3pt}

The new algorithm called \textit{SimpleDepthPose} follows a very simple concept with the following steps:

\begin{enumerate}[itemsep=-1pt, topsep=3pt, partopsep=0pt, parsep=3pt, labelindent=9pt]
  \item Predict joint coordinates for each color image
  \item Extract the distance of each detected visible joint from the aligned depth images
  \item Group the 3D pose proposals into persons from the last time-step, or create new ones if necessary
  \item Filter outliers in each proposal group
  \item Average remaining proposals to get the final 3D pose
\end{enumerate}

\begin{figure}[htbp]
  \centering
  \includegraphics[width=0.4\linewidth]{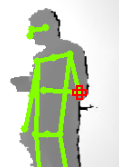}
  \caption{Visualization of the cross-shape used to extract the depth value for each joint in a zoom-in of the depth image. All pixels inside the cross are used to calculate the median depth distance.}
  \label{fig:sdp_cross}
\end{figure}

To predict the joint coordinates in step (1), basically every off-the-shelf pose estimator can be used. One important requirement is though, that the 2D pose estimation model is able to predict only directly visible keypoints, but no occluded ones, because they would result in extracting wrong depth values. 
Here a \textit{HigherHrNet}~\cite{cheng2020higherhrnet} model is used, trained and then finetuned on \textit{COCO}~\cite{lin2014microsoft} to predict only visible joints, and without the refinement step, because this was likely to add occluded joints again. A simpler grouping approach for the association scores was evaluated as well, which even resulted in overall better scores, but only if the refinement was kept, so it can not be used here.
\newline
\indent In step (2), the distance is extracted from the depth image by extracting the median value of pixels around the joint coordinates obtained from the RGB-based 2D pose estimator (see Figure~\ref{fig:sdp_cross}). The depth values are selected using two rectangular cutouts that form a cross-shape, to emphasize the center region and reduce outliers that would occur at the square's edges.
Afterward, a static per-joint offset is added to the distance, depending on the type of the joint, because a depth camera normally measures the distance to the surface, but the target location is the center of the joint.
These offsets are estimated using normal human proportions. For example, $3\,cm$ are added for shoulders and knees, or $1\,cm$ for the wrists. To account for larger persons or (thick) clothing those default values can be adapted.
At last, the poses are transformed from camera into world coordinates using the extrinsic calibrations.

In step (3), each 3D person proposal is then assigned to a person from the last time-step, by finding the closest match below a distance threshold. If no match is found, a new person is created. Old persons that were not matched are dropped after a certain number of frames.

Following this, a simple outlier removal step (4) is applied. The filter calculates the distance of each joint proposal to the averaged center of its neighboring joints, and if it is above a threshold, it is discarded.
In the case of a knee, for example, the neighbors are the hip and ankle. The idea is to remove proposals that are very far from the other joints, creating impossibly long limbs, which are likely to be wrong. Therefore the threshold (default $0.5m$) should cover all limb lengths of normal-sized persons ($<2m$).

\begin{figure}[htbp]
  \centering
  \begin{subfigure}{0.99\linewidth}
    \centering
    \includegraphics[width=0.99\linewidth]{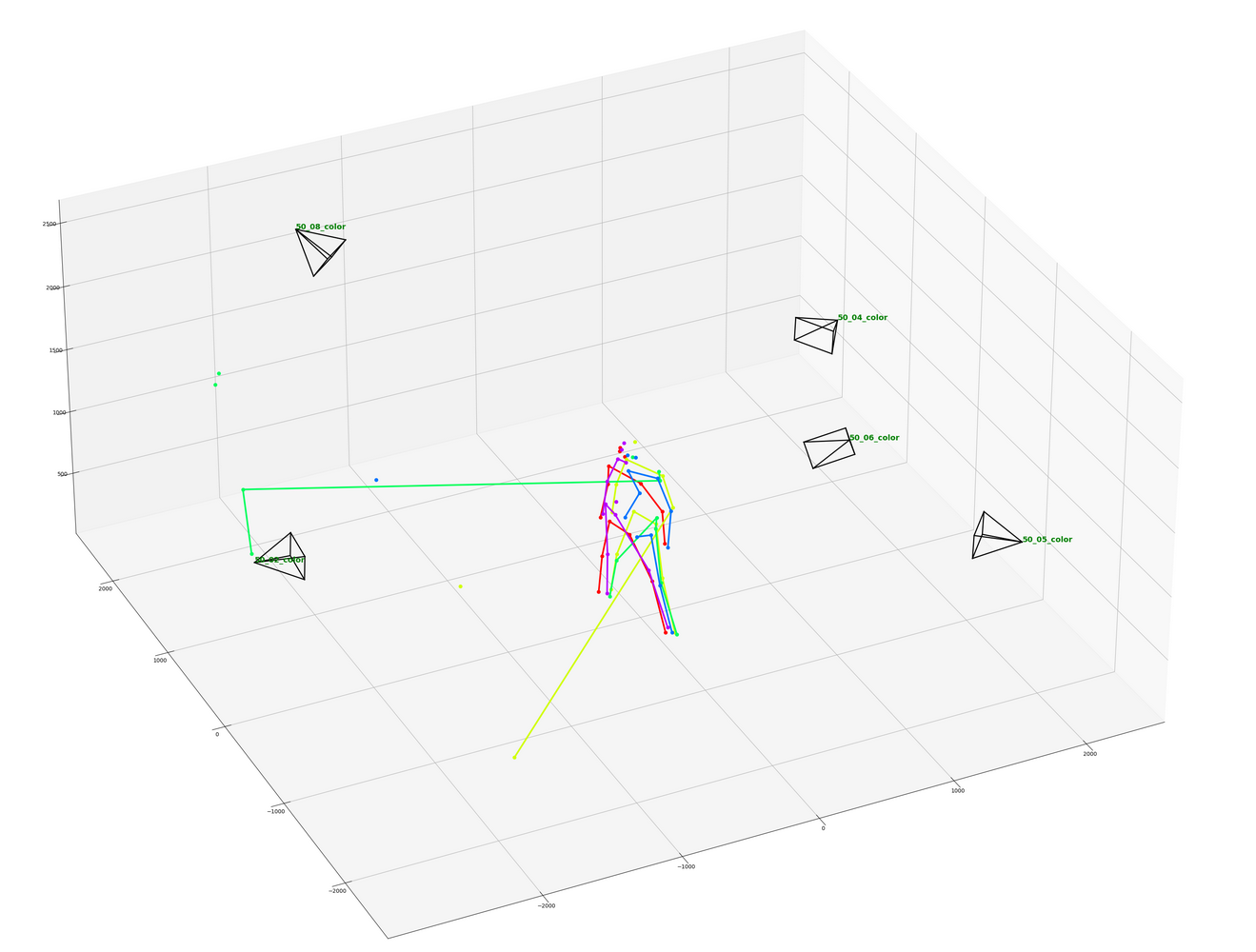}
  \end{subfigure}
  \hfill
  \begin{subfigure}{0.99\linewidth}
    \centering
    \includegraphics[width=0.59\linewidth]{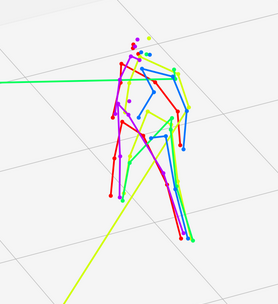}
  \end{subfigure}
  \hfill
  \begin{subfigure}{0.99\linewidth}
    \centering
    \includegraphics[width=0.99\linewidth]{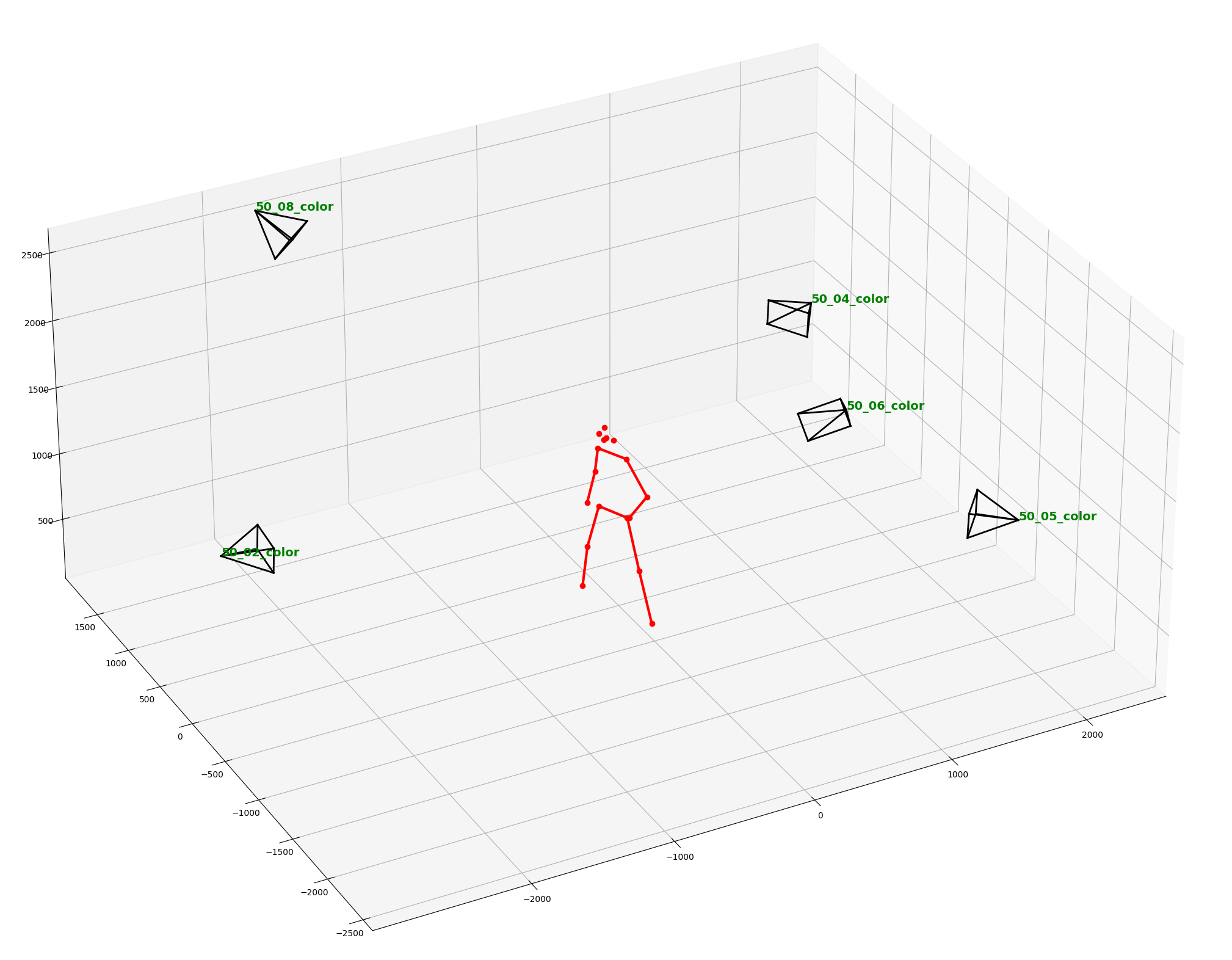}
  \end{subfigure}
  \caption{Example of the proposals for each view with some joint errors (top), a zoom-in on the per-view poses (center), and their fused result with the final joints (bottom).}
  \label{fig:sdp_fusion}
\end{figure}

Then, in the last step (5), if there are enough proposals for a joint, a center between them is calculated, and only the \textit{topk} (default $3$) closest proposals to this center are averaged into the new joint location. Wrong joint proposals are either caused by poor keypoint predictions or by errors in the depth image (especially at object edges), both resulting mostly in proposals far from the correct location of the joint. 
See Figure~\ref{fig:sdp_fusion} for an example of the proposals and their fused result.


\vspace{6pt}
In a direct comparison with \textit{OpenPTrack}, which is architecturally closest, \textit{SimpleDepthPose} is much simpler, but, as can be seen later, also more accurate. Two key differences are to account for the joint visibility and the improved concept of depth extraction. Another one is the filtering and merging approach. While \textit{OpenPTrack} uses a \textit{Kalman-Filter} to smooth the joint locations, which can lag behind in fast motions, \textit{SimpleDepthPose} only uses the joints from the last time-step, the other views, and the joint's direct neighbors to filter outlier joints before merging the remaining proposals to the final results.


\section{Dataset Evaluation}
\label{sec:ndgen1}
\vspace{-3pt}

In general, the number of multi-view multi-person depth datasets is very low, only two datasets were found that are suitable for this evaluation, \textit{mvor}~\cite{srivastav2018mvor} and \textit{panoptic}~\cite{joo2015panoptic}. Both datasets contain RGB and depth images, and the poses of the persons are labeled in 3D. 
The evaluation is performed using the same metrics as in \textit{VoxelKeypointFusion}~\cite{voxkeyfuse}, which evaluate the error of $13$ keypoints (2~shoulders, 2~hips, 2~elbows, 2~wrists, 2~knees, 2~ankles, 1~nose/head). The metrics are described in more detail in~\cite{voxkeyfuse}. The FPS was measured on a Nvidia-3090 as well.

\vspace{3pt}
\textit{Multi View Operation Room (MVOR)}~\cite{srivastav2018mvor} is more often used in literature and records an operation room from three different viewpoints. It is a relatively complicated dataset, since there is much occlusion, and the persons all have similar clothing. Only the upper body of a person is labeled, and most, but not all persons are labeled.

\begin{table*}[htp]
  \fontsize{7pt}{7pt}\selectfont
  \centering
  \begin{tabular}{@{}|l|c|cc|c|cc|c|c|c|@{}}
    \toprule
    Method \hspace{44pt}                                    & {\,}PCP{\,}   & \multicolumn{2}{c|}{PCK@100/500}           & {\,}MPJPE{\,} & \multicolumn{2}{c|}{Recall@100/500}        & {\,}Invalid{\,} & {}{\,}F1{\,}{} & {\,}FPS{\,}   \\
    \midrule
    MV3DReg \cite{kadkhodamohammadi2021generalizable}       & -             & \hspace{6pt} -             & -             & 176           & \hspace{6pt} -             & -             & -               & -              & -             \\
    VoxelPose                                               & 28.2          & \hspace{6pt} 10.8          & 35.9          & 119           & \hspace{6pt} 19.5          & 36.8          & 15.8            & 51.2           & 19.2          \\
    VoxelPose\,(synthetic)                                  & 36.3          & \hspace{6pt} 27.8          & 65.9          & 201           & \hspace{6pt} 15.2          & 72.4          & 76.7            & 35.3           & 8.3           \\
    Faster-VoxelPose                                        & 43.3          & \hspace{6pt} 31.1          & 55.2          & 120           & \hspace{6pt} 29.1          & 56.0          & 24.6            & 64.3           & 29.3          \\
    Faster-VoxelPose\,(synthetic)                           & 37.9          & \hspace{6pt} 28.1          & 45.5          & \textbf{109}  & \hspace{6pt} 29.4          & 46.1          & \textbf{7.7}    & 61.5           & \textbf{30.0} \\
    MvP                                                     & 0             & \hspace{6pt} 0             & 0.1           & 343           & \hspace{6pt} 0             & 0.1           & 99.9            & 0.1            & 8.8           \\
    PRGnet                                                  & 4.9           & \hspace{6pt} 3.6           & 6.2           & 120           & \hspace{6pt} 3.4           & 6.4           & 44.3            & 11.5           & 11.9          \\
    TEMPO                                                   & 10.4          & \hspace{6pt} 7.9           & 12.6          & 102           & \hspace{6pt} 8.8           & 12.7          & 14.0            & 22.2           & 20.3          \\
    SelfPose3d                                              & 48.8          & \hspace{6pt} 36.2          & 67.7          & 143           & \hspace{6pt} 31.1          & 70.2          & 36.5            & 66.7           & 13.0          \\
    mvpose                                                  & 45.9          & \hspace{6pt} 32.9          & 60.2          & 127           & \hspace{6pt} 27.1          & 61.3          & 18.5            & 70.0           & 0.8           \\
    mv3dpose                                                & 1.3           & \hspace{6pt} 0.7           & 2.8           & 235           & \hspace{6pt} 0.4           & 3.1           & 57.7            & 5.8            & 3.0           \\
    PartAwarePose                                           & 15.3          & \hspace{6pt} 10.3          & 25.5          & 201           & \hspace{6pt} 6.3           & 27.7          & 20.5            & 41.1           & 6.9           \\
    VoxelKeypointFusion                                     & \textbf{54.5} & \hspace{6pt} \textbf{43.9} & \textbf{75.1} & 128           & \hspace{6pt} \textbf{35.9} & \textbf{76.6} & 24.2            & \textbf{76.2}  & 11.3          \\
    \midrule
    \midrule
    MVDeep3DPS \cite{kadkhodamohammadi2017multi}            & -             & \hspace{6pt} -             & -             & 213           & \hspace{6pt} -             & -             & -               & -              & -             \\
    OpenPTrack                                              & 11.7          & \hspace{6pt} 9.9           & 26.8          & 323           & \hspace{6pt} 0.8           & 33.6          & 83.9            & 21.7           & 1.9           \\
    VoxelKeypointFusion                                     & 54.0          & \hspace{6pt} 44.2          & 72.2          & 119           & \hspace{6pt} 36.3          & 73.4          & 12.9            & 79.7           & 10.9          \\
    \midrule
    SimpleDepthPose                                         & \textbf{74.0} & \hspace{6pt} \textbf{62.0} & \textbf{94.3} & 113           & \hspace{6pt} \textbf{54.1} & \textbf{96.6} & 23.5            & \textbf{85.4}  & \textbf{37.2} \\
    \bottomrule
  \end{tabular}
  \caption{Transfer to \textit{mvor}~\cite{srivastav2018mvor} without and with depth. All other results without extra citations are taken from~\cite{voxkeyfuse}.}
  \label{tab:trans_mvor}
\end{table*}

Table~\ref{tab:trans_mvor} shows that many models have great problems with this setup, and only a few of them, including \textit{SimpleDepthPose}, reach a decent performance on this dataset. The problem is mainly caused by the many occlusions, which result in some persons, or most parts of them, being visible in only one image. All triangulation-based methods consequently struggle to detect such persons at all. When additionally using depth information, on the other hand, one view is enough to correctly detect them.

\begin{table*}[hbp]
\vspace{-9pt}
  \fontsize{7pt}{7pt}\selectfont
  \centering
  \begin{tabular}{@{}|l|c|cc|c|cc|c|c|c|@{}}
    \toprule
    Method \hspace{56pt} & {\,}PCP{\,}   & \multicolumn{2}{c|}{PCK@100/500}           & {\,}MPJPE{\,} & \multicolumn{2}{c|}{Recall@100/500}        & {\,}Invalid{\,} & {}{\,}F1{\,}{} & {\,}FPS{\,}   \\
    \midrule
    VoxelPose            & 98.5          & \hspace{6pt} 97.9          & 98.7          & 19.3          & \hspace{6pt} 98.7          & 98.7          & 1.1             & 98.8           & 8.0           \\
    Faster-VoxelPose     & 99.4          & \hspace{6pt} 98.6          & \textbf{99.9} & 20.5          & \hspace{6pt} 99.7          & \textbf{99.9} & \textbf{1.0}    & \textbf{99.5}  & \textbf{18.0} \\
    MvP                  & 97.6          & \hspace{6pt} 97.2          & 98.3          & 18.7          & \hspace{6pt} 98.0          & 98.5          & 15.8            & 90.8           & 8.9           \\
    PRGnet               & \textbf{99.5} & \hspace{6pt} \textbf{99.1} & \textbf{99.9} & 17.1          & \hspace{6pt} \textbf{99.9} & \textbf{99.9} & 2.0             & 99.0           & 6.8           \\
    TEMPO                & 98.1          & \hspace{6pt} 97.4          & 98.5          & \textbf{16.8} & \hspace{6pt} 98.4          & 98.4          & 2.4             & 98.0           & 5.1           \\
    SelfPose3d           & 99.3          & \hspace{6pt} 98.7          & 99.8          & 24.9          & \hspace{6pt} 99.7          & \textbf{99.9} & 8.0             & 95.7           & 7.1           \\
    \midrule
    \midrule
    mvpose               & 90.5          & \hspace{6pt} 75.9          & 97.5          & 83.6          & \hspace{6pt} 73.5          & 98.5          & 10.0            & 94.0           & 0.1           \\
    mv3dpose             & 84.5          & \hspace{6pt} 79.4          & 86.1          & 48.8          & \hspace{6pt} 81.8          & 86.4          & 15.6            & 85.4           & 1.3           \\
    PartAwarePose        & 89.8          & \hspace{6pt} 79.9          & 92.1          & 60.5          & \hspace{6pt} 83.1          & 93.0          & 1.4             & 95.8           & 1.5           \\
    VoxelKeypointFusion  & \textbf{97.1} & \hspace{6pt} \textbf{94.0} & \textbf{99.7} & \textbf{47.8} & \hspace{6pt} \textbf{97.3} & \textbf{99.9} & 2.4             & 98.7           & 4.2           \\
    \midrule
    OpenPTrack           & 83.0          & \hspace{6pt} 70.9          & 95.1          & 97.6          & \hspace{6pt} 68.9          & 97.2          & 15.5            & 90.4           & 1.8           \\
    VoxelKeypointFusion  & 92.6          & \hspace{6pt} 90.0          & 96.9          & 60.1          & \hspace{6pt} 85.4          & 97.8          & \textbf{0.1}    & \textbf{98.9}  & 4.0           \\
    SimpleDepthPose      & 96.9          & \hspace{6pt} 91.2          & \textbf{100}  & \textbf{45.5} & \hspace{6pt} \textbf{98.6} & \textbf{100}  & 4.7             & 97.6           & \textbf{17.7} \\
    \bottomrule
  \end{tabular}
  \caption{Replication of \textit{panoptic} results and transfer without and with depth. All other results are taken from~\cite{voxkeyfuse}.}
  \label{tab:res_panoptic}
\end{table*}

In this dataset it was notable that \textit{SimpleDepthPose} has large errors at hip joints, which on average are around $170\,mm$ off. In comparison, the upper body joints have an average error between $60\,mm$ to $100\,mm$. This is likely caused by the fact that the hip joints are much more often occluded, which sometimes leads to an incorrect assignment of depth values.
Other than that, it significantly outperforms all others in terms of the detected persons and keypoints. It is also faster than every other approach.

\vspace{3pt}
The second option for evaluation is the \textit{Panoptic}~\cite{joo2015panoptic} dataset, which is commonly used for evaluations of RGB-only approaches, but also contains depth recordings. The cameras are mounted in a dome-like structure and point to the center of it. The same evaluation approach as in \textit{VoxelKeypointFusion}~\cite{voxkeyfuse} was used here as well. Note that since the depth cameras were not time synchronized, their alignment to the color images and to the pose labels is not perfect. They were considered as belonging together if the time difference was below a threshold. 

As the results in table~\ref{tab:res_panoptic} show, the RGBD-based approaches outperform half of the algorithmic RGB-based ones in terms of the percentage of detected persons. \textit{SimpleDepthPose} shows a detection rate of persons and joints that is on the same level as the learned approaches that were trained in this setup, while being faster than most.
This is especially relevant in safety-critical applications, like in human-robot collaboration for example, where inaccurate joint or person estimations can be better handled, for example by generally increasing the required distances, than a missing one.
Some of the invalid predictions might be persons entering the room, which are often not labeled. \textit{SimpleDepthPose} already detects them if they are visible in one image, while many other approaches require at least two.


\vspace{-3pt}
\section{Ablation studies}
\label{sec:ablations}
\vspace{-3pt}

\begin{table*}[htp]
  \fontsize{7pt}{7pt}\selectfont
  \centering
  \begin{tabular}{@{}|l|c|cc|c|cc|c|c|c|@{}}
    \toprule
    Method \hspace{64pt}            & {\,}PCP{\,}   & \multicolumn{2}{c|}{PCK@100/500} & {\,}MPJPE{\,} & \multicolumn{2}{c|}{Recall@100/500}       & {\,}Invalid{\,} & {}{\,}F1{\,}{} & {\,}FPS{\,}  \\
    \midrule
    SDP\,(panoptic,\,with\,occluded\,kpts)    & 95.9 & \hspace{6pt} 90.3 & 99.8 & 49.1 & \hspace{6pt} 96.1 & 100  & 29.6 & 82.6 & 17.4 \\
    SDP\,(panoptic,\,without\,joint\,offsets) & 95.6 & \hspace{6pt} 88.8 & 100  & 52.9 & \hspace{6pt} 97.0 & 100  & 5.9 & 97.0 & 17.5 \\
    SDP\,(panoptic,\,cameras=1)               & 65.7 & \hspace{6pt} 58.6 & 84.2 & 155  & \hspace{6pt} 38.5 & 89.3 & 5.6  & 91.8 & 50.4 \\
    SDP\,(panoptic,\,cameras=3)               & 92.0 & \hspace{6pt} 82.5 & 99.5 & 64.3 & \hspace{6pt} 90.7 & 99.7 & 3.2  & 98.2 & 27.1 \\
    SDP\,(panoptic,\,cameras=10)              & 98.5 & \hspace{6pt} 96.8 & 99.9 & 37.5 & \hspace{6pt} 98.1 & 100  & 4.8  & 97.5 & 9.5  \\
    \midrule
    SDP\,(mvor,\,pc2vmap)                     & 63.7 & \hspace{6pt} 50.0 & 91.1 & 142  & \hspace{6pt} 29.5 & 94.7 & 21.2 & 86.0 & 3.1  \\
    SDP\,(mvor,\,pc2dimg)                     & 74.0 & \hspace{6pt} 61.7 & 94.5 & 114  & \hspace{6pt} 52.9 & 96.8 & 24.3 & 85.0 & 1.2  \\
    SDP\,(panoptic,\,pc2vmap)                 & 91.3 & \hspace{6pt} 82.3 & 99.4 & 70.4 & \hspace{6pt} 86.1 & 99.7 & 18.4 & 89.7 & 1.3  \\   
    SDP\,(panoptic,\,pc2dimg)                 & 96.6 & \hspace{6pt} 90.8 & 100  & 47.4 & \hspace{6pt} 97.7 & 100  & 7.9  & 95.9 & 0.5  \\
    \bottomrule
  \end{tabular}
  \caption{Ablation experiments with SimpleDepthPose.}
  \label{tab:abls}
\end{table*}

The visibility finetuning of \textit{SimpleDepthPose}, so that only directly visible joints are detected, has a relatively small impact on most metrics, but without it, the number of invalid predictions strongly increases. The added per-joint depth offsets notably improve the average position accuracy.

Even with only a single camera, the algorithm is able to detect most persons, even though the localization accuracy strongly decreases.
As expected, the results get better with more cameras, but the inference time increases as well. In case there are many cameras with overlapping views, implementing that persons need to be seen by multiple cameras to be valid could further reduce the number of invalid persons.

Due to the use of depth information from the cameras, the (learned) triangulation step can be skipped, and the algorithm is very fast. On \textit{Panoptic} the average time is about $2.6\,ms$ for the depth extraction and $0.4\,ms$ for multi-view fusion. For better performance, the fusion part is implemented in \textit{C++} and called through a \textit{Python} interface.

\vspace{3pt}
Besides directly pairing color and depth images, another option would be to fuse the depth information from all cameras first. For this all depth images are converted to point-clouds which are merged together. After that two different options were evaluated, the first one was to convert the point-cloud back to depth images again (\textit{pc2dimg}), and the second one was to convert it to a 3D~voxel-map (\textit{pc2vmap}, voxel resolution $5cm$). These concepts might be interesting if the depth information is not generated by depth cameras, but by other sensors instead, or if a point-cloud is already available, which is often the case in robotic applications. As can be seen in the results, both options output usable detections, with \textit{pc2dimg} being better. In comparison to the original approach of using the depth images directly without fusing them, the fusion takes some extra time (though the current implementation is not very efficient, so this could be faster), while the results are similar, so the fusion step is not considered necessary if depth images are available.


\section{Whole-body estimation}
\label{sec:wholebody}
\vspace{-3pt}

Similar to \textit{VoxelKeypointFusion} the algorithmic approach of \textit{SimpleDepthPose} can be easily extended to handle different input keypoints. This can for example be used to predict whole-body keypoints, which include additional face, foot, and finger keypoints.

While the 3D algorithm is easy to adjust, the 2D pose estimation part remains a challenge.
Extending the \textit{HigherHrNet} model did not work with its bottom-up concept, especially with the finger keypoints, since often only some of the visible fingers are labeled, and the default training process penalized predictions of unlabeled keypoints.
Instead, it is possible to use the whole-body keypoint model from \textit{RTMPose}~\cite{jiang2023rtmpose}, the same as in \textit{VoxelKeypointFusion}. Here only the visibility finetuning could not be included, because the face and hand keypoints do not contain information about their occlusion status.

This problem results in improvement possibilities for future works.
Since very often fingers are occluded by other fingers, they, as a result, have a poor localization performance. So currently the persons and all/most fingers and/or the face keypoints should be directly visible for whole-body estimations.


\section{Conclusion}
\label{sec:conclusion}
\vspace{-3pt}

This paper showed, through an evaluation of different datasets, that the proposed \textit{SimpleDepthPose} algorithm is a very fast and reliable approach if depth data is available, and also shows the best generalization results among other methods, without requiring any additional training.

One limitation is that since there is no neural refinement of the resulting 3D poses, the location accuracy is highly dependent on the accuracy of the depth images. Each keypoint also has to be visible in at least one image to be detected.

Following the results of the experiments, the use of depth data does not necessarily lead to more accurate results in terms of joint localization, but can significantly increase the number of detected keypoints and persons, especially in strongly occluded settings.
Since in many applications it is more important to detect persons and their poses at all, than to have a very accurate joint localization, using or integrating depth sensors can therefore be recommended, because as shown, they can increase the algorithm's performance.


\newpage

\renewcommand{\baselinestretch}{0.989}

{
    \small
    \bibliographystyle{ieeenat_fullname}
    \bibliography{mybib}
}

\end{document}